%% file: main.tex
\documentclass[
]{ceurart}
\usepackage{fancyvrb}
\usepackage[noabbrev,capitalize]{cleveref}

\usepackage[normalem]{ulem} 
\usepackage{xcolor}

\def\footurl#1{\footnote{\url{#1}}}

\usepackage{framed}

\def\tea{\textbf{Teacher}}  
\def\stu{\textbf{Student}}  
\def\eva{\textbf{Evaluator}}  

\def\model#1{\texttt{#1}}  
\def\ourgpt{\model{GPT\discretionary{-}{-}{-}4.1\discretionary{-}{-}{-}nano}}
\def\ourgptmini{\model{GPT\discretionary{-}{-}{-}4.1\discretionary{-}{-}{-}\textbf{mini}}}
\def\ourgptFULL{\model{GPT\discretionary{-}{-}{-}4.1\discretionary{-}{-}{-}nano\discretionary{-}{-}{-}2025-04-14}}
\def\ourgptminiFULL{\model{GPT\discretionary{-}{-}{-}4.1\discretionary{-}{-}{-}\textbf{mini}\discretionary{-}{-}{-}2025-04-14}}
\def\LLMindsTeacher{\model{DeepSeek plus LLaMA}}
\def\outstandingoutsidersFULL{\model{Gemma-3-27b-it} (by Outstanding Outsiders \cite{outstandingoutsiders})}
\def\outstandingoutsiders{\model{Gemma-3-27b-it}}
\def\LLMindsTeacherFULL{\LLMindsTeacher{} (\model{DeepSeek\discretionary{-}{-}{-}R1\discretionary{-}{-}{-}Distill\discretionary{-}{-}{-}Qwen-14b}, and \model{LLaMA-3.2-3b}, by the team LLMinds \cite{thatmakessense})}

\def\LLMindsStudent{\model{LLaMA plus DeepSeek} (\model{Llama-2-7b} plus \model{DeepSeek\discretionary{-}{-}{-}R1\discretionary{-}{-}{-}Distill\discretionary{-}{-}{-}Qwen\discretionary{-}{-}{-}1.5b}, by LLMinds \cite{thatmakessense})}

\def\BarFoo{\model{Mistral-7b}}
\def\BarFooFULL{\BarFoo{} (by the team BarFoo)}

\def\TeamUTK{\model{Llama-3-8b} (by TeamUTK)}
\usepackage{booktabs}
\usepackage{import}
\usepackage{amsmath}
\usepackage{placeins}
\usepackage{listings}
\begin{document}

\copyrightyear{2025}
\copyrightclause{Copyright for this paper by its authors.
  Use permitted under Creative Commons License Attribution 4.0
  International (CC BY 4.0).}

\conference{CLEF 2025 Working Notes, 9 -- 12 September 2025, Madrid, Spain}

\title{Overview of the {S}ensemaking {T}ask at the {ELOQUENT} 2025 {L}ab:
{LLMs} as {T}eachers, {S}tudents and {E}valuators}

\author[1]{Pavel Šindelář}
[%
email=sindelar@ufal.mff.cuni.cz,
]
\cormark[1]
\author[1]{Ondřej Bojar}[%
orcid=0000-0002-0606-0050,
email=bojar@ufal.mff.cuni.cz,
]
\cormark[1]

\address[1]{Charles University, Prague, Czech Republic}

\cortext[1]{Corresponding author.}

\begin{abstract}
ELOQUENT is a set of shared tasks that aims to create easily testable high-level criteria for evaluating generative language models. Sensemaking is one such shared task.

In Sensemaking, we try to assess how well generative models ``make sense out of a given text'' in three steps inspired by exams in a classroom setting: (1) Teacher systems should prepare a set of questions, (2) Student systems should answer these questions, and (3) Evaluator systems should score these answers, all adhering rather strictly to a given set of input materials.

We report on the 2025 edition of Sensemaking, where we had 7 sources of test
materials (fact-checking analyses of statements, textbooks, transcribed recordings of a lecture, and educational videos) spanning English, German, Ukrainian, and Czech languages.

This year, 4 teams participated, providing us with 2 Teacher submissions, 2 Student submissions, and 2 Evaluator submissions. We added baselines for Teacher and Student using commercial large language model systems. We devised a fully automatic evaluation procedure, which we compare to a minimalistic manual evaluation.

We were able to make some interesting observations. For the first task, the creation of questions, better evaluation strategies will still have to be devised because it is difficult to discern the quality of the various candidate question sets. In the second task, question answering, the LLMs examined overall perform acceptably, but restricting their answers to the given input texts remains problematic. In the third task, evaluation of question answers, our adversarial tests reveal that systems using the LLM-as-a-Judge paradigm erroneously rate both garbled question-answer pairs and answers to mixed-up questions as acceptable.
\end{abstract}


\maketitle

\section{Introduction}
Large language models (LLMs) have been shown to have a remarkable ability to summarize text, answer questions, and evaluate task performance. In the Sensemaking shared task run as part of the ELOQUENT Lab \cite{eloquent-2025-lncs} \footnote{\url{https://eloquent-lab.github.io/}} at CLEF 2025, we focus on a more constrained setting. We want to determine the capacity of an LLM-based system to ``make sense'' of a given material, adding yet another possible approach to machine reading or text comprehension. We assume, that the ability of a system to create questions that assess the understanding of arbitrary material is an indication that the system understands the material. We specify arbitrary here because many systems can create questions for material simply by reflecting on questions they have seen generated for similar material, which does not necessitate understanding the material, such generation is not possible for any arbitrary piece of material. The same reasoning as given above for question generation can be extended to the ability to answer arbitrary questions about materials and the ability to rate such answers given material. In general, our task could be attempted with a system using any generative language model. However, we correctly assumed that all teams would use LLMs in this first edition, so the evaluation is mostly focused on the capabilities and limitations of LLMs specifically.

Our experiments come from two domains: fact-check analysis of potentially disinformative texts and a reasonably broad range of educational topics. The educational domain is of particular interest because our evaluation rubric (pose questions, answer them, evaluate various responses) is very similar to what educators are regularly doing.
The use of generative language models in the domain of education is an area that has been studied extensively \cite{wang2023multiqg, farea2022evaluation, kamalloo2023evaluatingopendomainquestionanswering}. However, most of this research has focused on the ability of models to answer questions and finish assignments. Typically, models were not tested for the ability to exclusively rely on information from given material, but only for whether they were able to answer at all. We see such question answering as an insufficient method for evaluating the ability of a model to make sense of arbitrary material. For evaluation of abstractive question answering from context, there are not many procedures that do not rely on human-provided reference answers, and there is often no provision made to test whether the model is actually retrieving answers from the material rather than emitting its general knowledge. There is also a lack of evaluation procedures for the task of generating questions that comprehensively cover a given input text.

If we were creating an evaluation metric for educator tools or deploying a generative language model in the domain of education, we would need to cater to many requirements, such as adherence to curricula, ability to mimic the kind of questions and evaluations culturally expected, etc. For the sake of evaluating only basic material understanding as a starting point, we chose only very simple requirements on fluency, validity, adherence to context, and several other measures described in \cref{teacher-eval}. In the end, we did not have to test for fluency because most models used almost universally produced fluent text.

The paper is structured as follows: In \cref{sec:experimental}, we describe the organization of the shared task and baseline systems. \Cref{sec:data} provides a summary of the underlying texts. \Cref{teacher-eval,student-evaluation,evaluator-evaluation} provide the evaluation methodologies and results for the obtained \tea{}, \stu{}, and \eva{} systems, respectively. A brief discussion of our observations of the LLM-as-a-Judge usecase  across these tasks is provided in \cref{sec:observations}. We conclude in \cref{sec:conclusion} and list the limitations of this work in \cref{sec:limitations}.

\section{Experimental setup}
\label{sec:experimental}

The Sensemaking shared task in 2025 had three tracks, all voluntary:
\begin{enumerate}
\item \tea{} systems were given input materials organized into sections and were expected to generate quizzes as lists of questions for each such section of input material. In addition to each question, they were asked to provide reference answers, if possible, to aid in the evaluation. The baselines for the \tea{} track included questions generated, extracted from the original materials, or given by experts. Because such a baseline is not available for all materials, we also included baseline questions generated by a system using \ourgptFULL{}.
\item \stu{} systems were given input materials and quiz questions and were expected to provide answers to the questions (based on the input materials rather than general knowledge). The baselines for the \stu{} task included golden answers to questions extracted from the original materials (where available) and answers generated by a system using \ourgptFULL{}.
\item \eva{} systems worked with input materials, a question, and an answer, and were expected to score the answer on a scale of 0 to 100. We again used \ourgptFULL{} as an \eva{} baseline and also measured the extent to how the \eva{} systems correlate with .
\end{enumerate}

A total of four teams participated, each in various tasks. Three teams submitted valid submissions for two tasks, while one team focused only on \eva{} submission. Some submissions contained invalid output for too much of the dataset and were not included in the evaluation.

The data span 4 languages and 3 modalities. For the sake of the participants, all data were also provided in plaintext English form. Texts in languages other than English were machine-translated to English using the Google Cloud Translation - Basic (v2) system. We used the \model{whisper-turbo} model provided by the \model{openai-whisper 20240930} system for automatic speech recognition of speech input materials. All teams worked only with English data. Only one team chose to use the PDF modality in addition to plain text, using a system similar to \cite{wang2023multiqg}.

A very small subset of data was released as a development set prior to the evaluation period to illustrate the task to the participants: only one section for three kinds of materials. This was considered satisfactory since the only goal was for the participants to get an understanding of the kind of input their systems will need to handle. Providing such a small development set allowed us to have a large, mostly uncorrelated test set. There were no recommended training sets. Although we did some small-scale experiments with a few datasets, such as SQuAD \cite{squad}, we were unable to find any dataset that we were confident in recommending.

\section{Data}
\label{sec:data}
The material and questions were sourced from freely-accessible websites, YouTube, a university lecture, a study, and an academic book. The answers were acquired from one study and one non-copyrighted book website's administrators. To maintain copyright restrictions, we do not release our processed form of the dataset publicly, but we can grant access to it upon request, as we did for registered task participants.

There were a total of 7 data sources (referred to as material kinds in the paper). 
\begin{enumerate}
    \item A database of public statements by Czech politicians that were fact-checked by the Demagog.cz project. An example of a datapoint constructed from this database can be seen in Figure \ref{figsm}. (demagog-statements-public)
    \item A 19th-century book providing alternative logical and theological arguments about many natural phenomena (these views are not endorsed by the authors). (flat-earth-book)
    \item An English neural machine translation publication \cite{koehn2020neural}. (nmt-book)
    \item Audio from video recordings of a university lecture given by Ondřej Bojar. (nmt-class)
    \item A selection of various popular science material and European parliament speeches in German. The questions are in Czech and provided by \cite{javorsky-etal-2022-continuous}. (popular)
    \item Ukrainian high-school textbooks on biology (ukr-biology). These materials had many questions, and the individual sections were very long. However, we were unable to get hold of reference answers. (ukr-biology)
    \item English world history textbook from an open education resource (OER) website. For educators, reference questions and answers were also available. The textbook had many additional information subsections that were not related to the main body of information; these subsections had to be removed for the sake of brevity and text continuity. (world-history)
\end{enumerate}
For the \stu{} and \eva{} systems, we also constructed adversarial inputs from both the material and the other systems' outputs, as described in \cref{figsm2}.

\begin{figure}[th!]
\centering
\begin{framed}
     \begin{minipage}{0.92\textwidth} {\small
\import{./}{example-sensemaking.tex}}
\end{minipage}
\end{framed}

\caption{A sample test item (i.e., the input material and a question) derived from the Demagog database with answers and ratings by competing systems. We underlined several spans in the input material and in the question to highlight what the models should have ideally noticed in order to answer correctly. Only baseline GPT-4.1-nano managed this, so the automatic ratings by Gemma-3-27b are also mostly wrong.} \label{figsm}
\end{figure}

\begin{figure}[th!]
\centering
\begin{framed}
     \begin{minipage}{0.92\textwidth} {\small
\import{./}{example-sensemaking2.tex}}
\end{minipage}
\end{framed}
\caption{A sample adversarial test item (i.e., the material, and a misleading, unrelated question) derived from the Demagog database with answers and ratings by competing systems. The baseline GPT-4.1-nano succeeded in handling the adversarial input and was also correctly rated as perfect by Gemma-3-27b-it. Other models produce unrelated outputs, and in the first case (llama-3-8b-Instruct finetuned), this also confuses the rating system, which assigns 95\% to it.}
\label{figsm2}
\end{figure}

\FloatBarrier
\input{teacher_eval}
\FloatBarrier
\input{student_eval}
\FloatBarrier
\input{evaluator_evaluation}

\section{Our observations for the LLM-as-a-Judge usecase}
\label{sec:observations}

Across our experiments, we collected the first following observations, which will likely be relevant to anyone wishing to use LLMs for automatic scoring of various tasks, the so-called LLM-as-a-Judge setting. This strategy appeared twice in our shared task: (1) We used LLMs to evaluate our \tea{} submissions, and (2) our shared task participants relied on LLMs in their \eva{} systems.

Looking at \cref{adversarialanswers}, we can see that \eva{} systems sometimes fail to distinguish between correct answers and obviously wrong ones. On the other hand, considering the more thorough evaluation of \tea{} submissions in \cref{teacherratingexamination}, we can see that our rating of the \tea{} systems is very stable under change of LLM and prompt. From these two observations, it seems that the evaluation of the \tea{} system is significantly more stable than the evaluation of the \stu{} outputs by the \eva{} systems.\footnote{More tests of \tea{} evaluation would be needed for full confidence, because for now we could not do the same kind of adversarial tests with our \tea{} evaluation as we do with \eva{} systems.}
Here, we discuss the possible differences between the two settings and present them as recommendations for improving the LLM-as-a-Judge reliability. We nevertheless note that we have not yet conducted any experiments to verify these preliminary observations; we will likely do so in follow-up work.

There were many differences between our \tea{} system evaluation and the \eva{} system outputs. In \tea{} evaluation, we compare sets of questions instead of individual question-answer pairs rated in the \eva{} track. We also obviously only compare the text and the question when evaluating the \tea{} systems as opposed to the text, the question, and the answer. The increased stability may be entirely due to the different distributions of rated items in both evaluations. However, manual review showed that the quality of the \tea{} system outputs was, if anything, harder to distinguish. This should mean their rating should be more, not less, difficult. We identify the following factors as possible reasons why our \tea{} evaluation seems to be more reliable than \eva{}s performance:

\begin{itemize}
\item \textbf{Range of output values must be carefully selected.} 
In the \eva{} track, to give maximum flexibility to our contestants, we allowed the ratings to be in the range 0--100. When experimenting with the creation of the \eva{} baseline and evaluating the \eva{} submissions, we noticed that when prompted with just the 0--100 range, the systems often output only the values 0 and 100. When we designed our \eva{} baseline system, we instead chose to use 0--5 and multiply the result by 20. In the \tea{} systems, we only use the words ``least'' and ``most'' and only require the system to assign input question sets to statements to indicate rating, so the system does not have to interpret the meaning of numerical ratings.

\item{} \textbf{Answer correctness is conflated with semantic similarity.}
When we looked at how \eva{} systems rate nonsensical answers, we saw that when they were semantically similar, even almost nonsensical answers were often rated relatively highly. This is also corroborated by the inability to understand the specific requirements of a more complex question, as can be seen in \cref{figsm}.

\item{} \textbf{Judgements on text-level relations appear more stable.}
We think that part of the increased stability of the evaluation of the \tea{} system can be attributed to the fact that we only compare the text and the question as a whole. When evaluating the question answering from context, the LLM is required to consider both the meaning of the question in context and the correctness of the answer in context. In other words, evaluation of \tea{}s needs less semantic inference and shallow topic overlap is sufficient, whereas evaluation of \stu{}s needs to interpret both question and answer in the given context and assess their match.

\item{} \textbf{Relative judgements appear more stable.} 
Another factor that could cause increased stability in \tea{} evaluation is that our rating using \ourgpt{} was relative. This is not always possible, as we often want to rate a system in isolation and in absolute terms. We suspect that when processing the outputs of multiple systems at the same time, the LLM doing the rating can better understand its task, and thus become less prone to hallucination. We note that the relative rating could be emulated during the absolute evaluation using a few-shot method.

\end{itemize}

\section{Conclusions}
\label{sec:conclusion}

Systems that utilize LLMs can initially appear proficient in many tasks and seem to excel in text ``understanding''. In the Sensemaking task, we put this to a thorough, three-stage test. Participating systems and our baselines had to create questions to cover a given input material, answer them (based on the information in the material only) and evaluate such answers, again constrained by the input texts. This setting was inspired by the educational domain but applies to many other domains, including, e.g., fact-check analysis of potentially disinformative texts. As we have shown, most of the systems seemed to be similarly adept at handling all tested domains.

Our results indicate that, across the board, the models were sometimes able to generate and answer the questions. However, the results also show that they cannot be fully trusted to limit their knowledge to the materials provided.

The received \tea{} submissions and our baselines show that LLMs are reasonably good at converting text to questions about its content. The coverage can be easily ensured by extracting questions from every portion of the input. At the same time, the reference answers, if provided at all, were not very good, so the question construction appears to be done in a way detached from the actual content.

The answers of the \stu{} systems often did not make use of obvious clues about the answer in the materials. For example, in the fact-checking domain, the golden answer was spelled out explicitly in the text, even if a little hidden in a handful of unrelated statements, and yet the systems had difficulties extracting it.

The ratings produced by the \eva{} systems were unreliable and often were obviously wrong. This was striking, especially in our adversarial tests, where the \eva{} systems gave good scores to answers artificially damaged beyond comprehensibility, or to good answers connected to damaged questions. When the answers and questions were shuffled but generally coming from the same domain, the \eva{} systems were also fooled and gave relatively high scores. We see this result as an important caveat for the currently popular idea of using LLMs as judges.

Overall, our experiments with LLMs as judges seem to point to many imperfections and some promising potential paths. We hope to see more literature on the topic published and plan to do a thorough literature review in the future. 

The main conclusion of Sensemaking 2025 is that it is very difficult to achieve consistent results and progress without good automatic measures for evaluating text understanding systems in the examined tracks. The most reliable methods we used here were those that built upon adversarial inputs.

In this sense, our quest for ``sensemaking'' and its evaluation has only just begun. If LLMs indeed have the ability to understand text or validate if an agent exhibits such understanding, this capacity is far from easy to reach reliably. Most contestants in this task have used relatively simple methods and small models with a small or no amount of inference-time thinking, so there is still much experimentation to do.

\section{Limitations}
\label{sec:limitations}

Here we give a short summary of the limitations, including some already mentioned ones.
\begin{itemize}
\item Not counting our baselines, there were only two valid submissions for each track, barely enough to even have a comparison.
\item Because the number of dataset items increases exponentially as the tracks build on each other, we needed to prune the datasets in order to keep the compute requirements manageable for our contestants; this decreased the number and therefore the variety of the material sections used.
\item Submission items often have wildly different quality, so it is sometimes hard to compare some adversarial and non-adversarial entries when evaluating \eva{} systems.
\item When evaluating, we and the \eva{} systems focus only on the material section provided, because it is difficult to take into account the entire material at once; some \tea{} and \stu{} systems managed this and thus had a better chance. In future editions, it will likely be useful to ensure that the tracks are defined in a way that lends itself to easier evaluation.
\item In general, we ran Sensemaking 2025 in a rather exploratory way and the scope of this task thus did not allow us to focus precisely enough on the specific issues of each individual track. It might be beneficial to focus closely on only one of the tracks in the follow-up work and further editions of the task.
\end{itemize}

\begin{acknowledgments}
The authors acknowledge the support of the National Recovery Plan funded project MPO 60273/24/21300/21000 CEDMO 2.0 NPO and the funding from the Project OP JAK Mezisektorová spolupráce Nr. CZ.02.01.01/00/23\_020/0008518 named ``Jazykověda, umělá inteligence a jazykové a řečové technologie: od výzkumu k aplikacím.''
\end{acknowledgments}

\section*{Declaration on Generative AI}
While LLMs (as a particular type of generative AI tool) are the central element in this study (they create questions, answer them, and evaluate the answers), they were not used in the production of this text. The only exception is the sentence polishing feature built into Overleaf, where we sparingly accepted the suggested improvements.

\bibliography{citations}

\appendix
\section{Examining the stability of the \tea{} system evaluation}
\label{teacherratingexamination}
In \cref{tab:teacherLLM}, we can see that our \tea{} baseline using \ourgpt{} dominates the evaluation. One reason for this dominance could be the similarity between \ourgpt{} doing the \tea{} task and \ourgptmini{} evaluating it. The question we ask in this section is thus: Does the \tea{} evaluation change when we use different models or change the order of the requested entries in the prompt in \cref{teacherevalprompt}?

We chose to examine two alternative models for evaluation, \ourgpt{} and \model{Gemini\discretionary{-}{-}{-}2.5\discretionary{-}{-}{-}flash} as provided by the Google Gemini openAI compatibility API.\footurl{https://ai.google.dev/gemini-api/docs/openai} When we use the smaller one, i.e. \ourgpt{}, as seen in \cref{tab:teacherGPT}, we observe that the ratings of the \BarFoo{} system increase dramatically (1.48 to 1.83, far away from the minimum rank), and the ratings of all other systems seem to decrease relatively uniformly. The questions the \BarFoo{} system generated were often ranked worse in our manual review \cref{tab:teachermanual}, which indicates that the \ourgpt{} system is a worse judge than \ourgptmini{}. When we change the evaluation model to Gemini, we would expect that \ourgpt{} in the role of \tea{} loses some of its dominance. However, this does not happen, so we conclude that \ourgpt{} is maybe not the best model for \tea{} evaluation, but it was still the best \tea{} in our experiments; see \cref{tab:teacherGemini}.

In our second branch of examination, we check the stability of outputs when varying the prompt. For simplicity, we do not modify the actual wording of the prompt (\cref{teacherevalprompt}) but only change the order in which the evaluated outputs are presented to the \ourgptmini{} and \model{Gemini\discretionary{-}{-}{-}2.5\discretionary{-}{-}{-}flash} models.

Specifically, we experimented with swapping the pairs of ``least'' and ``most'' in the sequence, i.e. swapping the position and index of ENTRY1 with ENTRY2, ENTRY3 with ENTRY4, etc. The resulting ratings do not change much regardless if we stick to the \ourgpt{} model (\cref{tab:teacherGPTreversedpairs}) or use Gemini (\cref{tab:teacherGeminireversedpairs}).

Similarly, we do not see a large change even when we reverse the entire order of the entries in our prompt; see \cref{tab:teacherGPTreversed,tab:teacherGeminireversed}.

\begin{table}[h]
\caption{\tea{} submissions evaluated by \ourgpt{} (a different version than in \cref{tab:teacherLLM}). We report the average and standard deviation of the scores assigned to the provided sets of questions across all the tested materials.
For an easy comparison of \ourgptmini{} (used in \cref{tab:teacherLLM}) and \ourgpt{} (here), we also report the overall scores for each evaluated \tea{} across the categories in the last two columns of this table.
}
\label{tab:teacherGPT}
\centering
\small
\begin{tabular}{p{0.18\linewidth}p{0.12\linewidth}p{0.12\linewidth}p{0.12\linewidth}p{0.12\linewidth}|ll}
\toprule
type of evaluation & cover the material most & require thinking about different parts of the material & useful for learning the material & useful for learning to reason about the material & \multicolumn{2}{c}{Overall} \\
questions by &  &  &  &  & 4.1-mini & 4.1-nano \\
\midrule
DeepSeek plus LLaMA & 
2.18$\pm{}$0.82 & 
1.53$\pm{}$0.77 & 
1.76$\pm{}$0.74 & 
1.66$\pm{}$0.66 & 1.77$\pm{}$0.76 & 1.78$\pm{}$0.78 \\
Mistral-7b with BLIP & 
1.76$\pm{}$0.79 & 
1.88$\pm{}$0.63 & 
1.81$\pm{}$0.71 & 
1.88$\pm{}$0.77 & 1.48$\pm{}$0.67 & 1.83$\pm{}$0.72 \\
baseline from creators & 
1.71$\pm{}$0.77 & 
2.11$\pm{}$0.72 & 
1.86$\pm{}$0.80 & 
2.14$\pm{}$0.72 & 2.15$\pm{}$0.65 & 1.95$\pm{}$0.77 \\
baseline GPT-4.1-nano & \textbf{2.33$\pm{}$0.75} & \textbf{2.54$\pm{}$0.77} & \textbf{2.56$\pm{}$0.75} & \textbf{2.46$\pm{}$0.85} & \textbf{2.69$\pm{}$0.55} & \textbf{2.48$\pm{}$0.78} \\
\bottomrule
\end{tabular}
\end{table}

\begin{table}[h]
\caption{\tea{} submissions evaluated by \model{Gemini\discretionary{-}{-}{-}2.5\discretionary{-}{-}{-}flash}. We report the average and standard deviation of the scores assigned to the provided sets of questions across all the tested materials.
For an easy comparison of \ourgptmini{} (in \cref{tab:teacherLLM}) and \model{Gemini\discretionary{-}{-}{-}2.5\discretionary{-}{-}{-}flash}, we also report the overall scores for each evaluated \tea{} across the categories in the last two columns of this table.
}
\label{tab:teacherGemini}
\centering
\begin{tabular}{p{0.18\linewidth}p{0.12\linewidth}p{0.12\linewidth}p{0.12\linewidth}p{0.12\linewidth}|ll}
\toprule
type of evaluation & cover the material most & require thinking about different parts of the material & useful for learning the material & useful for learning to reason about the material & \multicolumn{2}{c}{Overall} \\
questions by &  &  &  &  & 4.1-mini & Gemini \\
\midrule
DeepSeek plus LLaMA & 1.96$\pm{}$0.73 & 1.82$\pm{}$0.69 & 2.04$\pm{}$0.81 & 1.88$\pm{}$0.75 & 1.77$\pm{}$0.76 & 1.93$\pm{}$0.74 \\
Mistral-7b with BLIP & 1.40$\pm{}$0.66 & 1.33$\pm{}$0.57 & 1.62$\pm{}$0.85 & 1.31$\pm{}$0.60 & 1.48$\pm{}$0.67 & 1.42$\pm{}$0.69 \\
baseline from creators & 2.19$\pm{}$0.79 & 2.14$\pm{}$0.76 & \textbf{2.31$\pm{}$0.82} & 2.06$\pm{}$0.58 & 2.15$\pm{}$0.65 & 2.17$\pm{}$0.74 \\
baseline GPT-4.1-nano & \textbf{2.52$\pm{}$0.67} & \textbf{2.75$\pm{}$0.48} & 2.19$\pm{}$0.66 & \textbf{2.75$\pm{}$0.56} & \textbf{2.69$\pm{}$0.55} & \textbf{2.55$\pm{}$0.63} \\
\bottomrule
\end{tabular}
\end{table}

\begin{table}[h]
\caption{Results using \ourgptmini{} with a similar methodology as in \cref{tab:teacherLLM} but with order of entries in the prompt (\cref{teacherevalprompt}) swapped within each least--most pair. The last two columns again compare the overall scores across categories to \ourgptmini{} with the original prompt.}
\label{tab:teacherGPTreversedpairs}
\centering
\begin{tabular}{p{0.18\linewidth}p{0.12\linewidth}p{0.12\linewidth}p{0.12\linewidth}p{0.12\linewidth}|ll}
\toprule
type of evaluation & cover the material most & require thinking about different parts of the material & useful for learning the material & useful for learning to reason about the material & \multicolumn{2}{c}{Overall} \\
             &  &  &  &  & orig & swapped \\
questions by &  &  &  &  & 4.1-mini & 4.1-mini \\
\midrule
DeepSeek plus LLaMA & 1.94$\pm{}$0.74 & 1.70$\pm{}$0.71 & \textbf{2.12$\pm{}$0.87} & 1.82$\pm{}$0.77 & 1.77$\pm{}$0.76 & 1.90$\pm{}$0.79 \\
Mistral-7b with BLIP & 1.81$\pm{}$0.83 & 1.55$\pm{}$0.67 & 2.02$\pm{}$0.92 & 1.55$\pm{}$0.67 & 1.48$\pm{}$0.67 & 1.73$\pm{}$0.80 \\
baseline from creators & 2.17$\pm{}$0.97 & 2.17$\pm{}$0.74 & 2.11$\pm{}$0.95 & 2.00$\pm{}$0.63 & 2.15$\pm{}$0.65 & 2.11$\pm{}$0.83 \\
baseline GPT-4.1-nano & \textbf{2.27$\pm{}$0.66} & \textbf{2.71$\pm{}$0.54} & 1.92$\pm{}$0.48 & \textbf{2.69$\pm{}$0.64} & \textbf{2.69$\pm{}$0.55} & \textbf{2.40$\pm{}$0.67} \\
\bottomrule
\end{tabular}
\end{table}

\begin{table}[h]
\caption{\model{Gemini\discretionary{-}{-}{-}2.5\discretionary{-}{-}{-}flash} with a similar methodology as in \cref{tab:teacherGemini} but with order of entries in the prompt (\cref{teacherevalprompt}) swapped within each least--most pair. The last two columns again compare the overall scores across categories to \ourgptmini{} with the original prompt.}
\label{tab:teacherGeminireversedpairs}
\centering
\begin{tabular}{p{0.18\linewidth}p{0.12\linewidth}p{0.12\linewidth}p{0.12\linewidth}p{0.12\linewidth}|ll}
\toprule
type of evaluation & cover the material most & require thinking about different parts of the material & useful for learning the material & useful for learning to reason about the material & \multicolumn{2}{c}{Overall} \\
             &  &  &  &  & orig & swapped \\
questions by &  &  &  &  & 4.1-mini & Gemini \\
\midrule
DeepSeek plus LLaMA & 2.06$\pm{}$0.79 & 2.00$\pm{}$0.64 & 2.16$\pm{}$0.82 & 2.00$\pm{}$0.70 & 1.77$\pm{}$0.76 & 2.06$\pm{}$0.74 \\
Mistral-7b with BLIP & 1.38$\pm{}$0.62 & 1.29$\pm{}$0.55 & 1.55$\pm{}$0.86 & 1.19$\pm{}$0.45 & 1.48$\pm{}$0.67 & 1.35$\pm{}$0.65 \\
baseline from creators & 2.11$\pm{}$0.75 & 2.00$\pm{}$0.76 & \textbf{2.36$\pm{}$0.87} & 2.11$\pm{}$0.57 & 2.15$\pm{}$0.65 & 2.15$\pm{}$0.75 \\
baseline GPT-4.1-nano & \textbf{2.52$\pm{}$0.64} & \textbf{2.73$\pm{}$0.56} & 2.10$\pm{}$0.50 & \textbf{2.73$\pm{}$0.60} & \textbf{2.69$\pm{}$0.55} & \textbf{2.52$\pm{}$0.63} \\
\bottomrule
\end{tabular}
\end{table}

\begin{table}[h]
\caption{Results using \ourgptmini{} with a similar methodology as \cref{tab:teacherLLM} but with order of entries in the prompt (\cref{teacherevalprompt}) reversed. The last two columns again compare the overall scores across categories to \ourgptmini{} with the original prompt.}
\label{tab:teacherGPTreversed}
\centering
\begin{tabular}{p{0.18\linewidth}p{0.12\linewidth}p{0.12\linewidth}p{0.12\linewidth}p{0.12\linewidth}|ll}
\toprule
type of evaluation & cover the material most & require thinking about different parts of the material & useful for learning the material & useful for learning to reason about the material & \multicolumn{2}{c}{Overall} \\
             &  &  &  &  & orig & reversed \\
questions by &  &  &  &  & 4.1-mini & 4.1-mini \\
\midrule
DeepSeek plus LLaMA & 2.04$\pm{}$0.78 & 2.04$\pm{}$0.86 & 1.98$\pm{}$0.80 & 2.04$\pm{}$0.78 & 1.77$\pm{}$0.76 & 2.02$\pm{}$0.80 \\
Mistral-7b with BLIP & 1.76$\pm{}$0.91 & 1.50$\pm{}$0.59 & 1.95$\pm{}$0.96 & 1.48$\pm{}$0.63 & 1.48$\pm{}$0.67 & 1.67$\pm{}$0.81 \\
baseline from creators & 2.14$\pm{}$0.87 & 1.83$\pm{}$0.65 & \textbf{2.14$\pm{}$0.83} & 1.81$\pm{}$0.67 & 2.15$\pm{}$0.65 & 1.98$\pm{}$0.77 \\
baseline GPT-4.1-nano & \textbf{2.21$\pm{}$0.67} & \textbf{2.65$\pm{}$0.59} & 2.10$\pm{}$0.69 & \textbf{2.69$\pm{}$0.58} & \textbf{2.69$\pm{}$0.55} & \textbf{2.41$\pm{}$0.68} \\
\bottomrule
\end{tabular}
\end{table}

\begin{table}[h]
\caption{
Results using \model{Gemini\discretionary{-}{-}{-}2.5\discretionary{-}{-}{-}flash} with a similar methodology as in \cref{tab:teacherGemini} but with the order of entries in the prompt (\cref{teacherevalprompt}) reversed. The last two columns again compare the overall scores across categories to \ourgptmini{} with the original prompt.}
\label{tab:teacherGeminireversed}
\centering
\begin{tabular}{p{0.18\linewidth}p{0.12\linewidth}p{0.12\linewidth}p{0.12\linewidth}p{0.12\linewidth}|ll}
\toprule
type of evaluation & cover the material most & require thinking about different parts of the material & useful for learning the material & useful for learning to reason about the material & \multicolumn{2}{c}{Overall} \\
             &  &  &  &  & orig & reversed \\
questions by &  &  &  &  & 4.1-mini & Gemini \\
\midrule
DeepSeek plus LLaMA & 2.14$\pm{}$0.76 & 1.98$\pm{}$0.77 & \textbf{2.24$\pm{}$0.69} & 1.98$\pm{}$0.68 & 1.77$\pm{}$0.76 & 2.08$\pm{}$0.73 \\
Mistral-7b with BLIP & 1.50$\pm{}$0.74 & 1.48$\pm{}$0.67 & 1.95$\pm{}$0.94 & 1.26$\pm{}$0.54 & 1.48$\pm{}$0.67 & 1.55$\pm{}$0.77 \\
baseline from creators & \textbf{2.33$\pm{}$0.86} & 2.22$\pm{}$0.76 & 2.19$\pm{}$0.89 & 2.06$\pm{}$0.71 & 2.15$\pm{}$0.65 & 2.20$\pm{}$0.81 \\
baseline GPT-4.1-nano & 2.19$\pm{}$0.69 & \textbf{2.44$\pm{}$0.73} & 1.65$\pm{}$0.62 & \textbf{2.73$\pm{}$0.53} & \textbf{2.69$\pm{}$0.55} & \textbf{2.25$\pm{}$0.75} \\
\bottomrule
\end{tabular}
\end{table}


\end{document}

%% file: example-sensemaking.tex
\textbf{Material section (truncated):}\,The former government ... introduced several \uline{programs to help} employees, businesses, and the self-employed (SVČ) during the Covid-19 pandemic. As for the current ... compensation programs, the \uline{Ministry of Labor and Social Affairs} ... includes , for example, ... the \uline{Antivirus} A and ... Antivirus B programs. The purpose of the first of these is ... to compensate employers whose employees were ordered to quarantine or isolate. Antivirus B was to ... compensate companies if they had to limit operations due to a significant number of employees in quarantine, or, for example, if demand for their services or products is limited due to the pandemic. For injured self-employed persons, but also for partners of small companies and people working on the DPČ and DPP, there is a ... Compensation Bonus program, which ... falls under the \uline{Ministry of Finance}. For ... example, the \uline{Ministry of Industry and Trade} was supposed to be responsible for ... the COVID 2021 subsidy program and the ... COVID Uncovered Costs program, ... intended for companies with a significant drop in sales. Although they were already prepared, the government ultimately decided ... We therefore assess Marian Jurečka... statement \uline{\uline{as true}}.\\
\textbf{Question automatically generated from database (truncated):}\,Determine whether the statement relevant to the given summary is TRUE or FALSE based on a summary and a number of statements. Only one of the following statements is relevant to the summary:\\
SECTION A: The mechanism (increase in pensions, note: Demagog.cz) is given by law (...), if since the last month taken for the ...\\
SECTION B: The proposal that I presented last year contained approximately 11, 11 measures in total. It also concerned widows, ...\\
SECTION C: So, the \uline{tools of assistance} (compensation for entrepreneurs, note: Demagog.cz) are shared here today mainly by 3 ministries, that is the \uline{Ministry of Labor and Social Affairs}, for us it is the programs of \uline{Antivirus} (... ), and then we also share the tools of assistance with the \uline{Ministry of Industry and Trade} and the \uline{Ministry of Finance}, where the other tools of assistance for entrepreneurs and tradesmen are. \\
 Respond only with either TRUE or FALSE and the correct section name, examples: TRUE, SECTION A.\\
\textbf{Reference answer:}\,TRUE, SECTION C\, \textbf{Rating by Gemma-3-27b:}\,100.00\\
\textbf{Answer by baseline GPT-4.1-nano:}\,TRUE, SECTION C\, \textbf{Rating by Gemma-3-27b:}\,100.00\\
\textbf{Answer by Llama-3-8b:}\,\textit<|assistant|> \textit{newline} FALSE \textit{newline} SECTION C.\,\textbf{Rating by Gemma-3-27b:}\,100.00\\
\textbf{Answer by Llama-2-7b plus DeepSeek-R1-Distill-Qwen-1.5b:}\,FALSE, SECTION C\, \textbf{Rating by Gemma-3-27b:}\, 100.00\\

%% file: example-sensemaking2.tex
\textbf{Material section (truncated)}:\,Pirate Party National  consists of an election committee and an election coordinator. The functioning of the election committee is regulated by  of the Election Code. The election committee currently has 6 people and its chairman is Senator ... . He was unanimously elected to the position by the members of the election committee at an introductory  held on July 19, 2019. Other members include  ...(representative of the media department),  ...(coordinator of the Liberec region and representative of the personnel department), ... ,  and  .... The election coordinator is  ... . The main ... of the election committee include approving the overall strategy and financing of the election campaign and coordinating the election staff. We therefore assess the statement as true.
\\\textbf{Question extracted from material (adversarial, because it is related to a different material)}\,How is human insulin produced using genetically modified bacteria?
\\\textbf{Answer by baseline GPT-4.1-nano:}\,The text does not provide information on how human insulin is produced using genetically modified bacteria. \textbf{Rating by: Gemma-3-27b-it}\,100.00
\\\textbf{Answer by llama-3-8b-Instruct finetuned:}\,<|assistant|>\textit{<newline>}The production process involves introducing a gene that codes for the human insulin precursor into a bacterium, such as E. coli, and then inducing the bacteria with a substance such as IPTG to express the gene and produce the insulin precursor. The insulin precursor is then processed into mature insulin using enzymes present in the bacterium. Finally, the mature insulin is harvested from the bacteria and purified using standard biotechnology procedures. \textbf{Rating by: Gemma-3-27b-it}\,95.00
\\\textbf{Answer by Llama-2-7b plus DeepSeek-R1-Distill-Qwen-1.5b:}\,The election committee approves the overall strategy and financing of the election campaign and coordinates the election staff. \textbf{Rating by: Gemma-3-27b-it}\,0.00

%% file: teacher_eval.tex
\section{\tea{} system evaluation}
\label{teacher-eval}
There were 2 fully valid submissions to the \tea{} task: \LLMindsTeacherFULL{} and \BarFooFULL{}. In total, 3 submissions were received, but the last was received late and contained only the \textit{<|assistant|>} token as the questions, so we disregarded it in the evaluation.

There were 2 baselines, the aforementioned expert-made questions and a system using \ourgpt{}. The \ourgpt{} baseline used a simple system, where we uniformly extracted spans of words from the section. We then prompted the LLM to generate one question for each span, answerable from that span; see \cref{teacherprompt}. In order to ensure that all the test set items received at least some questions, we also troubleshooted the baseline on the test set.
\label{unfair-troubleshooting}
This means it is artificially stronger when compared to the contestant systems. To ensure the output had the correct format, we used the OpenAI API's structured output functionality.

\begin{figure}[h]
\framebox{
\begin{minipage}{.95\textwidth}
\import{}{teacher_prompt.tex}
\end{minipage}}
\caption{The simple prompt used for the \tea{} baseline using \ourgpt{}. The prompt is relatively long because we wanted to make sure it did not fail on any part of the dataset. Further prompt optimization was surely possible.}
\label{teacherprompt}
\end{figure}

Evaluation of question generation is an underexplored and difficult task.
Generally, one would need to acquire many reference questions
\cite{oh2023evaluationquestiongenerationneeds} and compare them with the submitted questions with metrics such as ROUGE or
BLEU, or use human evaluators. We decided to use two methods of
evaluation: (1) a set of simple automatic methods, which should have a small
variance, and (2) a method relying on an LLM to hopefully handle the
semantics correctly. For the latter, we also explore a manually revised version of the LLM outputs on a small subset.

\subsection{Simple automatic evaluation}
\label{simple-autoeval}
The first approach to evaluation is designed to be as straightforward as possible, but
nevertheless considers several aspects of the generated questions: their
relevance, coverage of the source text, and mutual diversity of the questions.
All these quantities are derived from a common pre-processing of the input
material and the evaluated set of questions.

We start with a set of questions $Q$ and a text. First, we segment the text into a set of overlapping windows $W$. Then we obtain embeddings of $Q$ and $W$ using \model{paraphrase-multilingual-mpnet-base-v2}, a small sBERT-based \cite{sbert} model as provided by the \model{sentence-transformers 4.1.0} system.\footurl{https://sbert.net/} Using these embeddings and a similarity function, we define a relevance metric $r: Q\times W\rightarrow \mathbb{R}$. We measure the cosine similarity between the embeddings of each of the questions $q\in Q$ and each of the windows $w\in W$. Then, for each window, we recalculate the relevance to focus only on the most relevant questions. Cosine similarity can be defined as
\begin{equation}
\forall_{x \in \mathbb{R}^n,y \in \mathbb{R}^n} s_{\text{cosine}}(x,y) = \frac{x^Ty}{\sqrt{x^Tx}\sqrt{y^Ty}}.
\end{equation}
Let us have $c_x$ for each $x \in W$, a random variable that describes the similarities of embeddings of questions drawn from $Q$ with the window $x$.
Then we can express the relevance as
\begin{equation}
r(q,w) = \begin{cases}
s_{\text{cosine}}(M(q),M(w)) + \text{quantile}_{0.5}c_x &s_{\text{cosine}}(M(q),M(w)) > \text{quantile}_{0.95}c_x\\
0 &\text{otherwise}
\end{cases},
\end{equation}
where $M(q)$ and $M(w)$ represent the embeddings of the question and window, respectively.

We further normalize this relevance to get a categorical distribution with probabilities $p$ expressing the estimated likelihood that the question $q$ is related to the window $w$ of the text:
\begin{equation}
p: Q\times W\rightarrow \mathbb{R},  p(q,w) = \frac{r(q, w)}{\sum_{q',w'\in Q\times W} r(q',w')}
\end{equation}

Using this metric and distribution we measure the following three quantities.
\paragraph{Relevance}
Relevance measures how much a given set of questions and text relate to each other. It is aggregated from the relevance of a question to a window over the elements of $Q\times W$.

\begin{equation}
\text{relevance}(Q, W) = \sum_{q,w\in Q\times W} r(q,w)
\end{equation}

\paragraph{Coverage}
Coverage measures how uniformly the questions cover different parts of the text. Formally, we define it as the entropy of the distribution $m: W \rightarrow \mathtt{R}$, the marginal distribution on windows:
\begin{equation}
m(w) = \sum_{q \in Q} (p(q,w))  
\end{equation}

Maximizing the $H(m)$ is equivalent to making the coverage of the text by the questions as uniform as possible.

\paragraph{Diversity}
Diversity estimates how uniquely different questions cover different parts of the text. We compute it from the conditional distribution $p_q: W \rightarrow \mathbb{R}$ for a fixed $q\in Q$:

\begin{equation}
p_{q}(w) = \frac{p(q,w)}{\sum_{w’ \in W} p(q, w’)}
\end{equation}

Then the diversity is
\begin{equation}
\sum_{(q,q') \in Q\times Q}\text{KL-DIV}(p_{q} || p_{q'})
\end{equation}
High values of the Kullback-Leibler divergence indicate that the sequence of distributions $p_{q_1}, p_{q_2}, \dots , p_{q_{|Q|}}$ is diverse.

\paragraph{Evaluation procedure}
We observed that the ranges of the relevance, coverage and diversity values vary from document to document by orders of magnitude. Therefore, to obtain a final result, it would be problematic to take the average values over all documents. Instead, we decided to rank the systems on each document and report the average rank (with 4 denoting the best rank). The results can be seen in \cref{tab:pointsoverall,tab:pointsbyevaluation,tab:pointsbykind}. In cases where there were no questions submitted by the team, we automatically rate them as the worst with ties broken by chance.

We can see the variance of ratings over the different material sections is quite high. This could mean that the variance of this method was still quite high or that the systems performed rather irregularly in their question quality. We can also see that all the methods were rated very similarly. This is very unlikely to be a good estimate, as some of the systems scored significantly worse when rated manually, as will be shown below in \cref{manualsection}.

It is possible that if the evaluation method were better calibrated, e.g. by choosing a better way to convert the embedding pairs to a distribution, a better estimator could be created.
\FloatBarrier
\import{tables/}{teacher_automatic_all.tex}

\subsection{LLM evaluation}
For comparison with the simple automatic measure, we evaluate the questions by ranking unique triplets of submissions for a given document and rating them by \ourgptminiFULL{} in four categories.\footnote{Because the ranking model \ourgptmini{} was similar to the baseline model \ourgpt{}, we decided to also double-check the rankings using \model{Gemini\discretionary{-}{-}{-}2.5\discretionary{-}{-}{-}flash}; we got almost exactly the same results; see \cref{tab:teacherGemini} in the Appendix.}

We designed a complex prompt that asks the LLM to decide the worst and the best output from three submissions in multiple categories. The third evaluated system (i.e. the one not mentioned as the best or worst by the LLM) is implicitly ranked as the second one of the three. The highest (best) possible value of the rank is thus 3 in this evaluation.

In order to score all 4 systems (two submissions and two baselines), we repeatedly sample triplets from this set of 4 and average the obtained ranks.



The exact prompt is provided in \cref{teacherevalprompt}. The model should consider the situation of a student learning for an exam. Provided with the learning material and the three possible sets of questions, the model should go over 8 statements and for each of them indicate which of the three sets of questions matches the statement best. The statements are always paired; the first in a pair asks about the most covering/complex/etc. set of questions, the second asks about the least covering/complex/etc. set of questions.
We opted for this complex and structured request because we wanted to reuse the relatively long input material across the evaluated question sets. The comparative rating seemed to us to be easier for the model to process. Our limited experiments also indicated that this prompt led to more stable evaluations than other methods when comparing models.

\begin{figure}
\framebox{
\begin{minipage}{.95\textwidth}
You are required to fulfill an api request to a large language model, please respond with the requested filled in freeform text.\\
Simulate a randomly sampled university student required to prepare for an oral exam by studying the given material. The exam will require memorizing the material as well as learning to think about the topics in the material. The questions are not required to cover all the information in the material but must give at least some coverage of the main topics and pieces of information. We need you to fill the fields in our questionnaire about quiz questions so we can fill in our statistics.\\
You are a randomly sampled university student required to prepare for an oral exam by studying the given text, we need you to answer our questions about a quiz so we can fill in our statistics.\\
You are not sure what the exam will contain other than that it will be answerable using information from the text.
You are given quiz questions purely to help you understand the text better. Quiz questions are very different than those asked by the teacher during the oral exam.\\
You should be able answer these questions from the text provided.\\
Our scientific questionnaire is a simple format string. Each QUESTIONSET\_NUMBER: needs to be followed by the corresponding QUESTIONSET number.\\
\dots\\
(The text and numbered question sets in json format)\\
\dots\\
Please fill in the value QUESTIONSET\_NUMBER: \{\} with the correct QUESTIONSET numbers \\
ENTRY1: The questions require thinking about least different parts of the material. is most true about "QUESTIONSET\_NUMBER": \{\}\\
ENTRY2: The questions require thinking about most different parts of the material. is most true about "QUESTIONSET\_NUMBER": \{\}\\
ENTRY3: The questions cover the material least. is most true about "QUESTIONSET\_NUMBER": \{\}\\
ENTRY4: The questions cover the material most. is most true about "QUESTIONSET\_NUMBER": \{\}\\
ENTRY5: The questions are least useful for learning to reason about the material for the test. is most true about "QUESTIONSET\_NUMBER": \{\}\\
ENTRY6: The questions are most useful for learning to reason about the material for the test. is most true about "QUESTIONSET\_NUMBER": \{\}\\
ENTRY7: The questions are least useful for learning the material for the test. is most true about "QUESTIONSET\_NUMBER": \{\}\\
ENTRY8: The questions are most useful for learning the material for the test. is most true about "QUESTIONSET\_NUMBER": \{\}\\

\end{minipage}}
\caption{The prompt used in the \tea{} ranking. Some parts are repeated to
stress their importance. This prompt was constructed in a manual iterative
updates so it
is possible that there are  prompts easier to read that achieve a similar or
even better results.}
\label{teacherevalprompt}
\end{figure}

The results are provided  in \cref{tab:teacherLLM} and we can see that the outputs of our baseline system are scored the highest (remember the artificial advantage the baseline had, as discussed at the beginning of \cref{unfair-troubleshooting}), followed by \LLMindsTeacher{} submission by the LLMinds team. This is true broadly for all categories.

The main reason why the system using \BarFoo{} submitted by BarFoo is scored so low is that many of its quizzes contain questions related to a completely different part of the text. We suspect that the BarFoo team extended the context by concatenating the inputs but failed to filter the generated questions properly.

In contrast to automatic evaluation in \cref{simple-autoeval}, here we only score submissions on the texts where the participating systems provided some questions.

\paragraph{Manually revised LLM evaluation}
\label{manualsection}
In the final type of evaluation, we manually review the scores provided on a selected small subset of outputs. Our plan was to fix any clearly wrong assessments when skimming the provided sets of questions and referring back to the source material as needed. However, the sets of questions were very diverse and therefore difficult to compare. In the end, we left most of the automatic scores intact. We provide the manually revised ranks in \cref{tab:teachermanual}.

When reviewing the outputs, we observed some of the systems produced questions asking about the author or the title of the document, which is likely caused by a significant misunderstanding of the task.

\import{tables/}{teacher_LLM.tex}

\import{tables/}{teacher_manual.tex}



\subsection{Teacher evaluation overview}

We rate \LLMindsTeacher{} as the best of the 2 submissions. When looking at the manually revised ratings in \cref{tab:teachermanual} \LLMindsTeacher{} has a range of rankings, 2.1 to 2.5, that is similar to the range of rankings of the expert-made (manual) baseline and the \ourgpt{} baseline. The other submission, \BarFoo{}, has rankings in a significantly worse range of 1.4 to 1.5.

Overall, while the exact ratings of systems under these evaluation methods differ, we can see that a pattern emerges across all of the methods. The baseline, manually created by experts (the creators of the material), when available, typically scores among the better systems. It usually takes the second rank. In our experience, LLMs used as judges tended to rate other LLMs as better than human outputs, so we are not surprised to see this result. The fact that the expert-made questions from the creators score better in the LLM-based evaluation than \tea{}s' outputs can also be seen as a small sanity check of our evaluation methods, even if it did not last after our small manual revision.

The \LLMindsTeacher{} system seems to be only a little behind both baselines or even exceeds them at times, and does not fail on any material kind.

Based on our browsing of the data, but without providing any further details, we would like to mention that the systems that attempted to also provide reference answers generally provided very low-quality ones. We see this mismatch as surprising given the overall acceptability of the generated questions. Being able to derive questions from a given text and still be unable to answer them (based on the very same text) suggests that the models are probably using some shallow cues on question formation rather than processing the underlying meaning of the material. In the next section, we move to testing the ability to answer questions.

\FloatBarrier

%% file: teacher_prompt.tex
You are given a JSON dictionary labeled "Inputs" containing two fields:

- "Text": a passage of informational text
- "Words": a list of sentences or phrases extracted from the text

Your task is to generate a list of question-answer pairs based on this information. Your output should be similar to a teacher's assistant preparing material for deep comprehension and critical thinking. Each item in the list must be a JSON object with the following fields:
- "Question": a challenging, clear, and grammatically correct question
- "Answers": a list of five distinct example answers

Guidelines:  
1. Generate one question per item in the "Words" list.\\
2. Each question must be challenging, natural-sounding, and grammatically correct.\\  
3. Ensure diversity in the questions, with each one exploring a different angle or idea related to the text.\\
4. The questions should prompt critical thinking and help deepen understanding of the topic.\\
5. Do not refer to the words themselves as "phrases" or "items"—form questions directly about the concepts they describe.\\
6. Each "Answers" list must contain five distinct, well-reasoned, responses derived from the content of the text. These should demonstrate a mix of reasoning, inference, and interpretation, not just direct extraction. If five completely unique responses cannot be produced, then at least ensure the responses are phrased differently.\\
7. Each answer should make sense on its own and be completely independent of the other answers.\\
8. Avoid pronouns such as "you", "he", "she", "they", and "we", and instead use neutral terms like "one", "the author", or proper nouns where applicable.\\
9. Ensure each question stands on its own, so the topic and intent are clear without needing to see the full text.
10. Avoid questions that ask about the wording or phrasing in the passage.\\
11. Avoid referring to the speaker, narrator, or specific experiments in the text.\\
12. The output should be only a JSON array of {"Question": ..., "Answers": [...]} objects.\\
13. Use only English.\\

Example Input:
\begin{verbatim}
{
  "Text": "You are attacked countless times every day, ...",
  "Words": [
    "Your immune system is a complex army of billions of cells: ...",
    "Hundreds of millions of people in the West have been vaccinated ..."
  ]
} 
\end{verbatim}

Example Output:
\begin{verbatim}
[
  {
    "Question": "What roles do different immune cells play in ...",
    "Answers": [
      "Some immune cells function as soldiers, attacking and destroying harmful ...",
      "Some cells produce weapons against the infection ...",
      "Some components of the immune system serve as production centers ...",
      "Some components of the immune system serve as production ...",
      "Each type of immune cell contributes uniquely to ..."
    ]
  },
  ...
]
\end{verbatim}
Input:\\
\ldots\\
(The input formatted as above)\\
\ldots\\
\\

%% file: tables/teacher_automatic_all.tex
\begin{table}
\caption{Simple  automatic evaluation of \tea{} submissions: Mean rank with 4 being best overall.}
\label{tab:pointsoverall}
\centering
\begin{tabular}{ll}
\toprule
 & average rank \\
questions by &  \\
\midrule
DeepSeek plus LLaMA & 2.50$\pm{}$1.11 \\
Mistral-7b with BLIP & 2.48$\pm{}$1.13 \\
baseline from creators & 2.48$\pm{}$1.14 \\
baseline GPT-4.1-nano & 2.47$\pm{}$1.13 \\
\bottomrule
\end{tabular}
\end{table}
\begin{table}
\caption{Simple automatic evaluation of \tea{} submissions. Mean rank with 4 being best, grouped by evaluation measure: coverage, diversity and relevance.}
\label{tab:pointsbyevaluation}
\centering
\begin{tabular}{lll}
\toprule
 &  & average rank \\
questions by & type &  \\
\midrule
\multirow[t]{3}{*}{DeepSeek plus LLaMA} & coverage & 2.51$\pm{}$1.11 \\
 & diversity & 2.47$\pm{}$1.10 \\
 & relevance & 2.51$\pm{}$1.12 \\
\cline{1-3}
\multirow[t]{3}{*}{Mistral-7b with BLIP} & coverage & 2.44$\pm{}$1.13 \\
 & diversity & 2.51$\pm{}$1.13 \\
 & relevance & 2.50$\pm{}$1.13 \\
\cline{1-3}
\multirow[t]{3}{*}{baseline from creators} & coverage & 2.53$\pm{}$1.14 \\
 & diversity & 2.45$\pm{}$1.14 \\
 & relevance & 2.46$\pm{}$1.15 \\
\cline{1-3}
\multirow[t]{3}{*}{baseline GPT-4.1-nano} & coverage & 2.45$\pm{}$1.13 \\
 & diversity & 2.54$\pm{}$1.14 \\
 & relevance & 2.43$\pm{}$1.12 \\
\cline{1-3}
\bottomrule
\end{tabular}
\end{table}
\begin{table}
\caption{Simple automatic evaluation of \tea{} submissions: Mean rank with 4 being best, grouped by material kind. Best result for each material kind in bold.}\label{tab:pointsbykind}
\centering
\begin{tabular}{p{0.15\linewidth}p{0.15\linewidth}p{0.15\linewidth}p{0.15\linewidth}p{0.15\linewidth}}
\toprule
questions by & DeepSeek plus LLaMA & Mistral-7b with BLIP & baseline from creators & baseline GPT-4.1-nano \\
material kind &  &  &  &  \\
\midrule
demagog-statements-public & \textbf{2.51$\pm{}$0.98} & 2.41$\pm{}$1.06 & no questions & 2.02$\pm{}$0.92 \\
flat-earth-book & 2.64$\pm{}$1.19 & 2.55$\pm{}$0.99 & no questions & \textbf{2.86$\pm{}$1.00} \\
nmt-book & \textbf{2.85$\pm{}$0.99} & 2.70$\pm{}$1.07 & no questions & 2.41$\pm{}$0.93 \\
nmt-class & \textbf{2.49$\pm{}$1.12} & 2.48$\pm{}$1.13 & no questions & 2.47$\pm{}$1.14 \\
popular & \textbf{2.82$\pm{}$1.09} & 2.53$\pm{}$1.18 & 2.58$\pm{}$1.16 & 2.22$\pm{}$1.13 \\
ukr-biology & 2.40$\pm{}$1.12 & \textbf{2.69$\pm{}$1.12} & 2.58$\pm{}$1.12 & 2.57$\pm{}$1.09 \\
world-history & 2.46$\pm{}$1.11 & 2.31$\pm{}$1.33 & \textbf{2.87$\pm{}$1.08} & 2.76$\pm{}$1.06 \\
\bottomrule
\end{tabular}
\end{table}

%% file: tables/teacher_LLM.tex
\begin{table}
\caption{LLM-based evaluation of \tea{} submissions: Ranking in multiple categories as determined by  \ourgptminiFULL{} with the prompt provided in \cref{teacherevalprompt}.
We report the mean rank with 3 being best.}
\label{tab:teacherLLM}
\centering
\begin{tabular}{lp{0.15\linewidth}p{0.15\linewidth}p{0.15\linewidth}p{0.15\linewidth}}
\toprule
type of evaluation & cover the material well & require thinking about different parts of the material & useful for learning the material & useful for learning to reason about the material \\
questions by &  &  &  & \\
\midrule
DeepSeek plus LLaMA & 1.80$\pm{}$0.76 & 1.72$\pm{}$0.67 & 1.86$\pm{}$0.86 & 1.77$\pm{}$0.76 \\
Mistral-7b with BLIP & 1.52$\pm{}$0.67 & 1.36$\pm{}$0.58 & 1.43$\pm{}$0.70 & 1.83$\pm{}$0.72 \\
baseline from creators & 2.19$\pm{}$0.75 & 2.19$\pm{}$0.62 & 2.08$\pm{}$0.73 & 2.14$\pm{}$0.49 \\
baseline GPT-4.1-nano & \textbf{2.62$\pm{}$0.60} & \textbf{2.83$\pm{}$0.43} & \textbf{2.54$\pm{}$0.61} & \textbf{2.48$\pm{}$0.78}\\
\bottomrule
\end{tabular}
\end{table}

%% file: tables/teacher_manual.tex
\begin{table}
\caption{LLM-based evaluation of \tea{} submissions with manually revised ranks. We only revised the ordering for one document from each material kind. We changed about half of the individual rankings, most of them only by 1 point. We report the mean rank with 3 being the best.}
\label{tab:teachermanual}
\centering
\begin{tabular}{lp{0.15\linewidth}p{0.15\linewidth}p{0.15\linewidth}p{0.15\linewidth}}
\toprule
type of evaluation & cover the material well & require thinking about different parts of the material & useful for learning the material & useful for learning to reason about the material \\
questions by &  &  &  &  \\
\midrule
DeepSeek plus LLaMA & 2.27$\pm{}$0.70 & 2.25$\pm{}$0.87 & \textbf{2.40$\pm{}$0.63} & 2.27$\pm{}$0.70 \\
Mistral-7b with BLIP & 1.41$\pm{}$0.62 & 1.40$\pm{}$0.63 & 1.50$\pm{}$0.71 & 1.47$\pm{}$0.64 \\
baseline from creators & 2.00$\pm{}$0.87 & 1.62$\pm{}$0.74 & 2.00$\pm{}$0.87 & 1.78$\pm{}$0.83 \\
baseline GPT-4.1-nano & \textbf{2.58$\pm{}$0.67} & \textbf{2.47$\pm{}$0.61} & 2.36$\pm{}$0.81 & \textbf{2.50$\pm{}$0.76} \\
\bottomrule
\end{tabular}
\end{table}

%% file: student_eval.tex
\section{\stu{} system evaluation}
\label{student-evaluation}
For \stu{}, only two submissions were formatted in a recoverable way: \TeamUTK{} and
\LLMindsStudent{}. We again provided a simple baseline using \ourgpt{}, this time simply providing the entire section and a quiz, as can be seen in \cref{studentprompt}. To ensure that the output had the correct format, we again used the OpenAI API's structured output functionality.

\begin{figure}[h]
\framebox{
\begin{minipage}{.95\textwidth}
\import{}{student_prompt.tex}
\end{minipage}}
\caption{The simple prompt used for the \stu{} baseline using \ourgpt{}.}
\label{studentprompt}
\end{figure}

\import{tables/}{student.tex}

\subsection{Automatic evaluation}
We chose ROUGE-L Recall as the evaluation metric as advised by \cite{farea2022evaluation}. We preferred the Recall variant to ROUGE-L F1 because most of the answers are reasonably long, and we did not want to penalize longer answers. The results of these two metrics were very similar anyway. As can be seen in \cref{tab:scores,tab:scoresgold}, the results of this automatic evaluation of the systems on expert-made questions are much higher---up to almost twice as high for \LLMindsStudent{}, with 25 vs. 40. This indicates that the teacher-submitted questions were either harder to answer from the text provided, they were bad questions, or their reference answers were bad answers. Some submissions have relatively poor ratings, likely due to some responses having formatting issues or being based on questions/documents not matching the response indexation. Such issues should have been relatively easy to predict and debug on the devset.

The system \TeamUTK{} provided answers to only a smaller subset of the questions. The results of the systems on this subset can be seen in \cref{tab:scoresUTK}. They show that the TeamUTK model actually fares much better (up to 10 times) when we consider only this subset, but still underperforms the other models significantly.

\subsection{Correlations of ROUGE-L Recall scores between teams}

To verify our choice of the ROUGE-L Recall evaluation method, we tested whether expert-made questions could be reliably separated into easy and difficult questions from scores across the \stu{} systems. To this end, we calculate the Pearson correlation coefficients of ROUGE-L recall across all expert-made questions, for all pairs of system submissions.
We see that \model{Llama-3-8b} and \model{DeepSeek plus LLaMA} are rated significantly more similarly than any other pair, that is, 0.24 vs. 0.1 or -0.21. This makes sense as they use similar LLMs. This shows that for similar LLMs, the expert-made questions are similarly easy or hard.

On the other hand, it is strange that the ROUGE-L Recall scores of our GPT baseline and \LLMindsStudent{} are highly negatively correlated, that is, -0.31. One possible explanation is that each uses an approach to question answering that works for questions with different binary properties; for example, factual vs. abstractive. The different observed difficulty of expert-made questions would deserve further exploration.

\subsection{Manual evaluation}
\label{manualevaluation}
The answers were also manually evaluated, randomly choosing one set of questions
for each
data source. We excluded world history because for its sections, the set of all questions
was too large to be evaluated manually given our limited evaluation budget.

To maintain consistency in the evaluation procedure, we prepared simple
instructions in the form of a questionnaire; see
\cref{fig:studentevalquestionaire}. The questionnaire was manually filled in by the main task evaluator. The time was limited, so the examination per answer was very brief.

As we can see in \cref{tab:studentmanual}, the results of the manual evaluation seem to generally agree with the automatic evaluation. This might be mostly due to the fact that many answers have technical difficulties, which makes them quite easy to distinguish as incorrect. A more detailed breakdown can be seen in \cref{tab:studentmanualbycat}.

\import{tables/}{student_manual.tex}

\begin{figure}[h]
\framebox{
\begin{minipage}{.95\textwidth}
    For each question-answer pair, address each of these points on a single line numbered correspondingly, giving 1-5 on opinion-based points. Whenever we are talking about using the provided texts, this also means using elementary school-level background knowledge. If you would answer any point as 1, continue to the next question-answer pair:\\
    1. is correct.\\
    2. was created using only the provided texts.
\end{minipage}
}
    \caption{The manual evaluation questionnaire.}
    \label{fig:studentevalquestionaire}
\end{figure}

\subsection{Using question answering systems for fact-checking}
\label{demagog-questions-construction}
For determining the system's capacity to answer questions about fact-checking,
we derived a collection for question-answer pairs automatically
from the Demagog database, obtained
in May 2025.

The database contains manually selected and processed statements by
Czech politians and can be browsed online.\footurl{http://demagog.cz/}
Each item in the database is equipped with: (1) the original statement, i.e. a
very short excerpt, typically in the form of an abridged sentence, (2) a long
fact-check explanation, i.e. an elaborate analysis with many references to
public sources, carefully discussing the verity of the statement, (3) a short,
one-paragraph summary derived from the long explanation, (4) the final verdict
on the verity: true, untrue, misleading, impossible to decide. Importantly, the
verdict is typically explicitly stated at the end of the long summary in plain
Czech.

As input to the \stu{} task, we used three kinds of question-material combinations: 

\begin{enumerate}
	\item The materials are the combination of a statement, a short fact-check
explanation, and the corresponding long fact-check explanation. The questions are those
generated by \tea{} systems when given the material as input.
(demagog-statements-public-statement-w-explanation)
\item The materials are a single long fact-check explanation. The questions
are constructed so that each one contains a set of up to five possible statements
uttered by the same politician and labelled A--E in the question. At the end, the questions ask the student to determine
the statement related to the explanation provided in the material,  and to
report its truthfulness. As said, the ground-truth truthfulness is typically
present at the end of the explanation. The possible answers were
TRUE followed by SECTION (A-E), FALSE followed by SECTION (A-E), and if none of
the statements matched the explanation, we demanded the answer UNKNOWABLE.
(demagog-statements-public-determine-statement)
\item The materials are very short statements. The
questions are created by concatenating a set of up to five possible long fact-check explanations for statements from the same person, and then
ask the student to determine the relevant explanation and determine the
statement's truthfulness. As above, the explanation typically contains the verdict
spelled out in some plain language form.
The possible answers were TRUE followed by SECTION
(A-E), FALSE followed by SECTION (A-E), and if none of the statements matched the
explanation, we demanded the answer UNKNOWABLE.
(demagog-statements-public-determine-explanation)
\end{enumerate}

Converting a genuine fact-checking task to a multiple-choice question might not have been ideal, and in further iterations of this task we are going to experiment further with formulating the questions in a way that is understandable to both humans and even small LLMs.

In \cref{tab:scoresbykindgold}, we can see that \LLMindsStudent{} performed
significantly better than their overall average
on this task (44 vs. 71, and 69 vs. 40 in \cref{tab:scoresgold}).
This can be attributed in part to the fact that the verdict is present in the explanation and that the correspondence between the statement and the explanation
can often be gathered from the names they both mentioned. Both of these effects
were illustrated in \cref{figsm}. 

We attempted to make the questions more challenging by creating
the wrong options using data from the same speaker, but this was not enough, as
the statement and fact-check often contain other specific names of ministries or
politicians, and the content-bearing words also easily identify the topic.
In light of this, it is somewhat interesting that the
systems did not achieve a ROUGE-L Recall closer to 100.

In order to provide some quantitative details on the submitted systems' ability to extract readily available information from a little obfuscated text, we look at two different divisions of the ``demagog-statements-public-determine-statement'' subset:

\begin{enumerate}

\item We calculate vocabulary similarity using the Jaccard index over word forms (number of word forms occurring in both texts divided by the number of word forms occurring in either text) between the correct statement and the explanation. We then separate items where the correct statement has a similarity at least 1.5 times higher than any of the confounding statements in the item. We expect that word-level overlap is going to help LLMs in identifying the relevant statement, and indeed, 
we get a very
large difference in the accuracy of predicting the correct statement: 0.44 without the overlap
vs 0.83 with the high vocabulary overlap. While this could be caused by semantic differences in the groups, another
possible explanation is that the models are using the similarity in terms and
names and other shallow expressions instead of understanding the texts.

\item We look at all items in the ``demagog-statements-public-determine-explanation'' subset where the relevant explanation explicitly states the truthfulness with phrases ``as true'' or ``as
untrue'' and compare them to those where these phrases are not stated. In our data, all statements containing these phrases are rated as true or as untrue according to them. We evaluated the accuracy of predicting truthfulness in each of these two groups separately. While the correct truthfulness could have been read directly from these phrases, so the former group of items should have obtained much higher accuracy than the latter one, we measured rather similar accuracies, 0.58 for the former and 0.41 for the latter.
This suggests that the
systems were unable to understand that these particular words in the conclusion of the explanation are a solid indicator.

\end{enumerate}

\subsection{\stu{} evaluation overview}
\label{studenoverview}
It seems that even for the questions where \TeamUTK{} submitted the answers,
\LLMindsStudent{} provides better results. Both sets of submissions appear to score worse than our \ourgpt{} baseline.

However, the reliability of LLMs on \stu{}-like systems  is very far from acceptable. Our small analysis on the fact-checking obfuscated multiple choice questions indicates that LLMs use shallow word overlap to identify relevant sections in the input, and are not able to understand which words from the input are directly providing the expected answer, even if they are present.

%% file: student_prompt.tex
You are a knowledgeable assistant.
Your task is to answer a set of questions using ONLY the information explicitly provided in the following text.
Make sure to follow the following rules to the letter.

1. Do NOT use any outside knowledge or make assumptions.\\
2. Do not explicitly reference the text.\\
3. Avoid pronouns such as "you", "he", "she", "they", and "we", and instead use neutral terms like "one", "the author", or proper nouns where applicable.\\
4. Use only English.\\
5. Avoid referring to the speaker, narrator, or specific experiments in the text.\\
6. For each question, return an object with the following fields:\\

- "answer": If the question is answerable, provide the exact direct answer in the form requested by the question, do not describe your answer unless the question asks you to. If the question is not exactly answerable from the text try to answer anyway to the best of your ability.\\
- "answerable": A boolean indicating whether the question can be directly answered using the text.\\
7. Return exactly as many answers as there are items in the question list.\\
8. Only say a question is not answerable if the text does not provide ANY information relevant to the question, otherwise answer the question.\\

Format your responses as a list of such objects, one per question.

Example response format:
\begin{verbatim}
[
  {{
    "answer": "The immune soldiers usually take care of attacks ...",
    "answerable": true
  }},
  {{
    "answer": "The monastery includes a baroque library, farm ...",
    "answerable": true
  }},
  {{
    "answer": "The text mentions that it is difficult to ...",
    "answerable": false
  }}
]
\end{verbatim}

Reference text:
\ldots\\
(The material section)
\ldots\\
Questions:
\ldots\\
(The list of questions)
\ldots\\

%% file: tables/student.tex
\begin{table}[t]
\caption{The average ROUGE-L Recall scores of \stu{} submissions on all questions.}
\label{tab:scores}
\begin{tabular}{ll}
\toprule
 & rating \\
answered by &  \\
\midrule
Llama plus DeepSeek & 25$\pm{}$33 \\
baseline GPT-4.1-nano & \textbf{49$\pm{}$34} \\
Llama-3-8b & 02$\pm{}$11 \\
\bottomrule
\end{tabular}
\end{table}

\begin{table}[t]
\caption{The average ROUGE-L Recall scores of \stu{} submissions on questions where all teams submitted.}
\label{tab:scoresUTK}
\begin{tabular}{ll}
\toprule
 & rating \\
answered by &  \\
\midrule
Llama plus DeepSeek & 32$\pm{}$36 \\
baseline GPT-4.1-nano & \textbf{57$\pm{}$38} \\
Llama-3-8b & 19$\pm{}$25 \\
\bottomrule
\end{tabular}
\end{table}

\begin{table}[t]
\caption{The average ROUGE-L Recall scores of \stu{} submissions on expert-made questions.}
\label{tab:scoresgold}
\begin{tabular}{ll}
\toprule
 & rating \\
answered by &  \\
\midrule
Llama plus DeepSeek & 40$\pm{}$41 \\
baseline GPT-4.1-nano & \textbf{57$\pm{}$41} \\
Llama-3-8b & 04$\pm{}$15 \\
\bottomrule
\end{tabular}
\end{table}

\begin{table}[t]
\caption{The average ROUGE-L Recall scores of \stu{} submissions on expert-made questions grouped by kind.}
\label{tab:scoresbykindgold}
\centering
\begin{tabular}{p{0.2\textwidth}p{0.12\textwidth}p{0.12\textwidth}p{0.12\textwidth}p{0.12\textwidth}p{0.12\textwidth}}
\toprule
material kind & demagog-statements-public-determine-explanation & demagog-statements-public-determine-statement & demagog-statements-public-statement-w\-explanation & popular \\
answered by &  &  &  &  &  \\
\midrule
Llama plus DeepSeek& 44$\pm{}$42 & \textbf{71$\pm{}$28} & \textbf{69$\pm{}$39} & \textbf{25$\pm{}$12} \\
baseline GPT-4.1-nano & \textbf{51$\pm{}$39} & 46$\pm{}$36 & 26$\pm{}$34 & 08$\pm{}$04 \\
Llama-3-8b & 11$\pm{}$23 & 00$\pm{}$00 & 00$\pm{}$00 & no anwers \\
\bottomrule
\end{tabular}
\end{table}

\begin{table}[t]
\caption{For each pair of submissions we report the Pearson correlation coefficient of ROUGE-L Recall scores on expert-made questions.}
\label{tab:crosscorrelationgold}
\begin{tabular}{p{0.2\linewidth}p{0.2\linewidth}p{0.2\linewidth}p{0.2\linewidth}}
\toprule
 & Llama-2-7b plus DeepSeek-R1-Distill-Qwen-1.5b & baseline GPT-4.1-nano & Llama-3-8b \\
\midrule
Llama plus DeepSeek & 1.00 & -0.31 & 0.24 \\
baseline GPT-4.1-nano & -0.31 & 1.00 & 0.1 \\
Llama-3-8b & 0.24 & 0.1 & 1.00 \\
\bottomrule
\end{tabular}
\end{table}

%% file: tables/student_manual.tex
\begin{table}
\caption{The manual Likert ratings 1-5, 5 is best.}
\label{tab:studentmanual}
\centering
\begin{tabular}{lll}
\toprule
rating kind $\rightarrow$ & answer is correct & answer is based solely on the material \\
answered by $\downarrow$&  &  \\
\midrule
Llama plus DeepSeek & 2.78$\pm{}$1.31 & 2.97$\pm{}$1.45 \\
baseline GPT-4.1-nano & \textbf{3.35$\pm{}$1.25} & \textbf{3.58$\pm{}$1.44} \\
Llama-3-8b & 2.00$\pm{}$1.41 & 2.20$\pm{}$1.70 \\
\bottomrule
\end{tabular}
\end{table}

\begin{table}
\caption{The manual Likert ratings 1-5, 5 is best, by data category.}
\label{tab:studentmanualbycat}
\centering
\begin{tabular}{p{0.18\linewidth}p{0.18\linewidth}p{0.18\linewidth}p{0.18\linewidth}}
\toprule
 & rating kind & is correct & was created from the material \\
answered by & material kind &  &  \\
\midrule
Llama plus DeepSeek & demagog-statements-public & 2.49$\pm{}$1.29 & 2.49$\pm{}$1.29 \\
 & flat-earth-book & 1.60$\pm{}$0.85 & 1.80$\pm{}$1.13 \\
 & nmt-book & 2.05$\pm{}$1.48 & 1.98$\pm{}$1.38 \\
 & nmt-class & 3.80$\pm{}$1.13 & \textbf{4.60$\pm{}$0.57} \\
 & popular & \textbf{4.47$\pm{}$0.19} & 4.67$\pm{}$0.47 \\
 & ukr-biology & 2.29$\pm{}$0.75 & 2.29$\pm{}$0.75 \\
\hline
baseline GPT-4.1-nano & demagog-statements-public & \textbf{3.24$\pm{}$0.06} & \textbf{3.24$\pm{}$0.06} \\
 & flat-earth-book & \textbf{2.80$\pm{}$2.55} & \textbf{3.00$\pm{}$2.83} \\
 & nmt-book & \textbf{3.22$\pm{}$1.46} & \textbf{3.64$\pm{}$1.93} \\
 & nmt-class & \textbf{3.90$\pm{}$0.99} & 4.20$\pm{}$1.13 \\
 & popular & 4.38$\pm{}$0.31 & \textbf{4.83$\pm{}$0.24} \\
 & ukr-biology & \textbf{2.58$\pm{}$1.72} & \textbf{2.58$\pm{}$1.72} \\
\hline
llama-3-8b & flat-earth-book & 2.00$\pm{}$1.41 & 2.20$\pm{}$1.70 \\
\bottomrule
\end{tabular}
\end{table}

%% file: evaluator_evaluation.tex
\section{\eva{} system evaluation}
\label{evaluator-evaluation}
There were 2 valid \eva{} submissions, \outstandingoutsidersFULL{} and \BarFoo{}. We decided to limit our evaluation to responses in which both teams submitted an evaluation. This did not change the results as the system with more submitted ratings still leads by a large margin.

We again provided a baseline using \ourgpt{} to compare with the submissions of the contestants. In this baseline system, instead of using the full scale of 0--100 in the LLM prompt, we use a scale of 0--5 so that we can describe the meaning of each rating in words, as can be seen in \cref{evaluatorprompt}.

\begin{figure}[h]
\framebox{
\begin{minipage}{.95\textwidth}
\import{}{evaluator_prompt.tex}
\end{minipage}}
\caption{The simple prompt used for the \eva{} baseline using \ourgpt{}.}
\label{evaluatorprompt}
\end{figure}

The \eva{} systems sometimes get confused and use a scale from 0-1. This is obvious from the fact that they never do such fine-grained ratings at any other part of the 0-100 scale and the fact that they sometimes provide real-valued ratings instead of the integer-valued ratings requested. We took the liberty of multiplying all ratings less than or equal to 1 by 100 to recover as much usable data as possible.

\subsection{Adversarial answers}
\label{adversarialanswers}
The most interesting direct way to check validity is simply to provide the
system with a tailored adversarial input. One possible way to create such an
input is to take an answer that is related to the material section as closely as
possible, but not to the question itself. As we can see in
\cref{tab:adversarialanswersswapped}, most \eva{} systems, as expected, rate these inputs as
worse responses, but the decrease is not nearly as drastic as we would have
liked. Since these answers are almost always completely unrelated to the
questions, the average rating should fall close to 0.

\eva{} systems show similar ignorance when we shuffle words in the answer or swap the material section from which the question and answer originated. As can be seen in
\cref{tab:adversarialanswers}, we also see a relatively large decrease, but not
values close to 0.

\subsection{Adversarial questions using \stu{} submissions}
\label{adv-questions-using-stu-outputs}
Another simple way to check the validity of an \eva{} system is to test it on
adversarial questions. These were created by shuffling words, replacing them with random text, or swapping questions between different kinds of material. As we can see in
\cref{tab:adversarialquestions}, most systems rate the answers to these
questions as worse, for example, the baseline reduced its ratings from 65 to 49. However, the decrease is not nearly as drastic as we might have
hoped, given that such questions are almost always completely nonsensical. Some
of this could be caused by some \stu{} systems detecting that they are given
unanswerable questions and adapting their response, as can be seen in \cref{figsm}, with \eva{} systems scoring such responses high. However, that is unlikely to entirely account for a decrease with such a small magnitude.

\subsection{Scoring human-made answers to human-made questions}
\label{scoring-human-answers-to-human-questions}
Finally, we can also assess how \eva{} systems score golden
answers to expert-made questions.

In \cref{tab:golditems}, we see that on average the golden answer to a golden question receives a higher score than the \stu{} answer to the same question, the expert-made questions get up to 94 percent average rating for \outstandingoutsiders{}.

\cref{tab:datacategory} below shows that one of the systems, \outstandingoutsiders{}, actually gives a mean score of 100 to the golden answers for documents in the category demagog-determine-explanation, which means that it rated them all with the most fitting rating.

As we have pointed out in \cref{tab:golditems} earlier, the other system, Mistral, at least provides golden answers with higher scores than the student answers, even if the average rating increase from 62 for non-gold to 72 for gold gives only a 10-point difference --- a significantly lower difference than the variance of its scores given to golden answers to different questions.

Interestingly enough, our \eva{} baseline using \ourgpt{} produced counter-intuitive outputs for one of the Demagog-derived question sets. It graded the golden answers in ``demagog-statements-public-determine-explanation'' significantly lower than the \stu{} answers; 45.00 vs 77.78, which is almost a 23-point difference, as seen in the detailed \cref{tab:datacategory}.

\import{tables/}{evaluator-additional_with_baseline.tex}

\subsection{Agreement between different \eva{} systems}

One way of checking the overall stability of the automatic evaluation provided by \eva{} systems is to test how often they agree in their output scores for a given text-question-answer tuple.
In general, for all pairs of \eva{} systems, when both systems gave a valid rating, these ratings often agreed with each other. They agreed in up to 90\% of the cases, as illustrated in \cref{tab:accuracies}. As expected, they most often agreed when rating the correct golden answers.

It is interesting to note that when dealing with adversarial input, the \eva{} systems disagreed more often than when rating non adversarial inputs. They got agreement percentages of around 60 on this adversarial input as opposed to percentages of around 70 on golden and \stu{} inputs. We see this observation as a possible basis for improving the reliability of the systems: a decrease in agreement by two or more independent systems could be seen as an indicator that the item is problematic and its scoring should be decreased, or is completely inappropriate. However, this is still not an ideal technique for the evaluated \eva{} systems as there is at least one clear counterexample: The agreement of \eva{} systems when they are evaluating answers generated  by \stu{}s to questions with shuffled words is almost as high and sometimes even higher than when rating \stu{} system answers to non-adversarial questions; see the last lines in the lower vs. upper part of \cref{tab:accuracies}.

\begin{table}
\caption{Agreement in evaluations by different \eva{} systems. In this table, we report the mean of the indicator function $I_{f_1f_2}(x) = 1 \iff c(f_1(x)) = c(f_2(x))$ for each pair of systems $f_1,f_2$ and the way in which the question and answers were created/modified. The class $c(y)$ of a rating $y$ is determined via quantizing $y$ into bins defined by ranges, class 1 corresponds to scores of [0-33), class 2 to scores of [33-66), and class 3 to scores of [66-100]. The upper part of the table uses genuine question-answer pairs, the lower part uses different adversaries.}
\label{tab:accuracies}
\centering
\begin{tabular}{lp{0.14\textwidth}p{0.14\textwidth}p{0.14\textwidth}}
\toprule
adversarial category & Gemma-3-27b-it $\sim$ Mistral-7b & Gemma-3-27b-it $\sim$ baseline GPT-4.1-nano & Mistral-7b $\sim$ baseline GPT-4.1-nano \\
\midrule
golden answer & 0.76 & 0.90 & 0.76 \\
answer by \stu{} system & 0.68 & 0.82 & 0.68 \\
\midrule
answers swapped & 0.43 & 0.65 & 0.47 \\
answer random text & 0.55 & 0.57 & 0.57 \\
question random text & 0.56 & 0.32 & 0.58 \\
words in answer shuffled & 0.57 & 0.56 & 0.47 \\
words in question shuffled & 0.69 & 0.85 & 0.56 \\
\bottomrule
\end{tabular}
\end{table}

\subsection{Comparison with metrics using reference answers}
A more indirect way is to compare the results of the \eva{} system with
the metrics we used to rate \stu{} systems in \cref{student-evaluation}.

Instead of directly correlating ROUGE scores with \eva{} predictions, we are curious to assess \eva{}s' ability to predict the rough correctness in absolute terms. We thus divide ROUGE-L and \eva{} scores into three classes, equidistantly covering the full range of 0--100. Then we determine how accurately the \eva{} scores fall into the same class as the ROUGE-L score and average this across all question-answer pairs and all \stu{} submissions.

In \cref{tab:correlations}, we see that the overall accuracy for non-adversarial questions is around 0.33 for all systems. This means that it is about as good as randomly guessing because there are only 3 classes. This could be primarily attributed to the low precision, below 0.1, in predicting when ROUGE-L Recall scores an answer as half correct (denoted ``class 2'' in the tables). This makes sense as it is for these sorts of answer where any metrics on question answering quality disagree most. 

As can be seen in \cref{tab:correlationsgold}, the models lack the ability to correctly score the answers with the lowest ROUGE-L Recall (denoted ``class 1'' in the tables) for expert-made questions. For most systems the accuracy of agreement with ROUGE-L Recall is higher in the case when dealing with expert-made questions, that is 0.19 increase for \outstandingoutsiders{} and 0.13 increase for \ourgpt{} baseline. In \cref{tab:correlationsadversarial}, we can see that when dealing with \stu{} answers to adversarial questions the systems seem to agree with ROUGE-L slightly more often than overall, even though it should be easy to match the usually low scores given to them. It should also be noted that some \stu{} answers to adversarial questions do achieve a high rating purely because of the inadequacies of ROUGE-L Recall.


\subsection{Using \eva{} systems for checking fact-check explanation validity}
As we can see in \cref{tab:correlationsstatement}, when predicting the rating of the \stu{} answers which aim to identify the related statement and ``read off' its verity (see \cref{demagog-questions-construction} for the description of this data subset), \outstandingoutsiders{} performs much better than Mistral, esp. in the high ROUGE-L range and performs on par with the baseline using \ourgpt{}.

\import{tables/}{evaluator_with_baseline.tex}

\subsection{Broader comparison of \eva{} outputs with our evaluation results}
\label{broader-comparison-of-eva-and-ours}

The left column of \cref{tab:answeringsystem} presents the mean
\eva{} scores assigned to the answers of the respective \stu{} submissions. These scores suggest that the \LLMindsStudent{} \stu{} system delivers the best answers, followed by \TeamUTK{} and finally by the GPT baseline.

When we compare this ranking to the results presented in \stu{} evaluation (\cref{tab:scores,tab:studentmanual}, discussed in \cref{studenoverview}), we see that the order of contestants given by the mean rating of the \eva{} systems is different from the order that ROUGE-L Recall and manual evaluation gave us. In particular, \cref{studenoverview} concludes that the GPT baseline was better than both \LLMindsStudent{} and \TeamUTK{}.

This significant disagreement between the manual ratings and ROUGE-L Recall on the one hand and \eva{} average ratings on the other hand may indicate that the \eva{} system ratings are incorrect. This likely was not caused by any undue preference because of similar LLMs, because all \eva{} systems used models of different strains than those used in \stu{} systems.

A more detailed analysis, inspecting the evaluations down to the level of domains and likely also individual questions and answers, would be necessary for a full explanation of this discrepancy. Our tentative conclusion is that \eva{} ratings are generally unreliable.

\begin{table}
\caption{We report two values for each participating team. The left column reports the mean rating by all participating \eva{}s of the given team's \stu{} system answers (excluding adversarial items). The right column reports the mean rating by all participating \eva{}s of reference answers by the team's \tea{} system to the questions this \tea{} system generated. Bear in mind that all teams used different prompts or even models or approaches for their \tea{} and \stu{} submissions.}
\label{tab:answeringsystem}
\begin{tabular}{lll}
\toprule
answers by & in the \stu{} mode & in the \tea{} mode \\
\midrule
BarFoo & not applicable & 70.97$\pm{}$36.74 \\
reference answers extracted from material & not applicable & 86.62$\pm{}$30.03 \\
baseline GPT & 73.04$\pm{}$37.18  & 64.43$\pm{}$31.96 \\
TeamUTK & 68.23$\pm{}$44.42 & not applicable \\
LLMinds & 68.04$\pm{}$39.56  & 70.09$\pm{}$37.87 \\
\bottomrule
\end{tabular}
\end{table}

\subsection{\tea{} vs \stu{} system ability to answer questions}
\label{tea-vs-stu-answering-ability}

\tea{} task participants were asked to provide not just questions, but also reference answers, whenever possible. Not all participants did this, but when they did and when they also submitted a \stu{} system, we can examine whether the answers made by the \tea{}s are better than the answers made by the separate \stu{} systems.
We present this comparison in \cref{tab:answeringsystem}.

We can see that for the LLMinds team, the reference answers by their \tea{} system are rated relatively closely behind the answers by their \stu{} system. This is quite strange because for the reference answer, the 
LLMinds \tea{} returned just an untreated small excerpt of the original material, which served as the basis for creating the question. Such excerpts are not phrased as an answer and do not point out the specific piece of information requested, even though they almost always contain all the relevant information. This suggests that \eva{}s' scores reflect some form of information overlap but are not very good at checking if the provided answer is actually in a form appropriate for answering the question.

For our GPT baseline, the \eva{} ratings of the answers constructed in the \tea{} mode were, as expected, higher than the ratings of the answers provided in the \stu{} mode. This could be explained by the fact that, in the \tea{} mode, the model knew that the questions were answerable from the text because it generated them like that. It also probably helped that the baseline \tea{} was prompted with the instruction that the answer to the generated question should be contained in the text. However, the fact that these reference answers are rated worse than the golden answers to expert-made questions again indicates some inadequacy on the part of the \eva{} systems or on the part of the GPT \tea{} baseline.

\subsection{\eva{} evaluation overview}
Based on the accuracy in predicting the ROUGE-L Recall scores (\cref{tab:correlations}) and considering the general match between the ROUGE-L Recall score and manual rating, as presented in \cref{manualevaluation}, it seems that the
\eva{} submission using \model{Gemma-3-27b-it} is the best. The other validation approaches and a
limited manual qualitative review seem to corroborate this fact.

All \eva{} systems,
however, seem to be very far from perfect. In particular, they are very easily fooled by
adversarial inputs (\cref{adversarialanswers,adv-questions-using-stu-outputs}), such as nonsensical questions or answers. What is even
harder to fix in the long run is their low reliability when the evaluated questions
and answers are all on the same topic, but shuffled. In this setting, the tested \eva{}
systems assign relatively high scores even to mismatching
question-answer pairs.

Our further analyses also support our concern that evaluation systems that use currently available LLMs may be unreliable. 
For example, the discrepancy between \eva{} outputs and ROUGE-L Recall (which matches the manually assigned scores) in \cref{broader-comparison-of-eva-and-ours}, or the high acceptance of direct citations of the original text snippets instead of proper answers to questions in \cref{tea-vs-stu-answering-ability}.
As another example, we give the occasional failures, even for the best performing \eva{}s, such as the counter-intuitive results of \ourgpt{} in \cref{scoring-human-answers-to-human-questions}.

%% file: evaluator_prompt.tex
Please respond with the requested plain text.\\
You are given a list of question answer pairs. You are required to rate the quality of the answer on a scale from 0 to 5.\\
The scale is as follows: 0 means the answer is completely wrong, 1 means the answer is partially useful, 2 means the answer is useful, 3 means the answer is very useful, 4 means the answer is almost correct, and 5 means the answer is completely correct.

Fill in a format strings with the ratings integers as per this EXAMPLE:

"RATINGS:\\
rating of ANSWER 1: \{\}"

OUTPUT:\\
"rating of ANSWER 1: 0"

REFERENCE TEXT:\\
\ldots\\
(The material section)\\
\ldots\\

QUESTION-ANSWER:\\
\ldots\\
(The question answer pairs stored as a JSON dictionary.)\\
\ldots\\

Fill in the following format string with the ratings integers

"RATINGS:\\
rating of ANSWER 1: \{\}\\
\ldots"

OUTPUT:

%% file: tables/evaluator-additional_with_baseline.tex
\begin{table}
\caption{Adversarial test of \eva{}s by swapping the question for a different one from
the same document while keeping the answer intact.
The left column contains the ratings
for unaltered items, the right columns contains the ratings with questions that were replaced by random ones or ones from a different kind of material or with words shuffled.
The scores for questions in the right column should have been substantially lower. The underlying set of materials is fixed.
}
\label{tab:adversarialanswersswapped}
\centering
\begin{tabular}{lll}
\toprule
 & rating non adversarial & rating adversarial \\
rating by &  &  \\
\midrule
Gemma-3-27b-it & 82.90$\pm{}$27.49 & 59.44$\pm{}$43.00 \\
Mistral-7b & 62.47$\pm{}$47.86 & 57.65$\pm{}$49.37 \\
baseline GPT-4.1-nano & 65.54$\pm{}$37.02 & 26.38$\pm{}$35.84 \\
\bottomrule
\end{tabular}
\end{table}

\begin{table}
\caption{Adversarial tests for \eva{}s by shuffling words in the answer or swapping in a material of a different kind.
The left column contains the ratings
for unaltered items, the right columns contains the ratings with
answers where the scores should have been substantially lower. The underlying set of questions corresponding to the answers is fixed.}
\label{tab:adversarialanswers}
\centering
\begin{tabular}{lll}
\toprule
 & rating non adversarial & rating adversarial \\
rating by &  &  \\
\midrule
Gemma-3-27b-it & 82.90$\pm{}$27.49 & 32.86$\pm{}$40.27 \\
Mistral-7b & 62.47$\pm{}$47.86 & 55.37$\pm{}$49.60 \\
baseline GPT-4.1-nano & 65.54$\pm{}$37.02 & 13.33$\pm{}$27.05 \\
\bottomrule
\end{tabular}
\end{table}

\begin{table}
\caption{Adversarial tests for \eva{}s by replacing the answer with an answer to a different question from the same section if possible; if it is not possible we use an answer to a question from the same material kind.}
\label{tab:adversarialquestions}
\centering
\begin{tabular}{lll}
\toprule
 & rating non adversarial & rating adversarial \\
rating by &  &  \\
\midrule
Gemma-3-27b-it & 82.90$\pm{}$27.49 & 67.40$\pm{}$40.42 \\
Mistral-7b & 62.47$\pm{}$47.86 & 58.80$\pm{}$48.72 \\
baseline GPT-4.1-nano & 65.54$\pm{}$37.02 & 49.38$\pm{}$42.67 \\
\bottomrule
\end{tabular}
\end{table}

\begin{table}
\caption{The comparison of how \eva{}s rate the expert-made answers. The left column contains ratings for answers to expert-made questions generated by \stu{} submissions, the right column contains ratings of expert-made answers to expert-made questions. Adversarial entries are not included.}
\label{tab:golditems}
\centering
\begin{tabular}{lll}
\toprule
 & rating non-gold & rating gold \\
rating by &  &  \\
\midrule
Gemma-3-27b-it & 81.50$\pm{}$28.26 & 94.69$\pm{}$15.68 \\
Mistral-7b & 62.15$\pm{}$47.92 & 72.73$\pm{}$46.71 \\
baseline GPT-4.1-nano & 63.71$\pm{}$36.82 & 80.94$\pm{}$35.35 \\
\bottomrule
\end{tabular}
\end{table}

%% file: tables/evaluator_with_baseline.tex
\begin{table}
\caption{
The mean ratings of \eva{} systems on questions automatically generated from the Demagog.cz database as described in \cref{demagog-questions-construction}.}
\label{tab:datacategory}
\begin{tabular}{p{0.18\linewidth}p{0.18\linewidth}p{0.18\linewidth}p{0.18\linewidth}
p{0.18\linewidth}}
\toprule
 & rating by $\rightarrow$& Gemma-3-27b-it & Mistral-7b & baseline GPT-4.1-nano\\
material kind & answers made by $\downarrow$ &  &  \\
\midrule
demagog-statements-public-determine-explanation & experts & 100.00$\pm{}$0.00 & No ratings & 45.00$\pm{}$48.70 \\
 & \stu{} systems & 66.56$\pm{}$36.30 & 89.63$\pm{}$31.06 & 77.78$\pm{}$36.87\\
\hline
demagog-statements-public-determine-statement & experts & 93.72$\pm{}$17.89 & 57.14$\pm{}$53.45 & 84.26$\pm{}$32.55 \\
 & \stu{} systems & 77.35$\pm{}$32.37 & 75.80$\pm{}$42.69 & 64.44$\pm{}$43.61 \\
 \hline
demagog-statements-public-statement-w-explanation & experts & 87.02$\pm{}$21.44 & 63.90$\pm{}$47.88 & 65.32$\pm{}$30.08 \\
 & \stu{} systems & 80.19$\pm{}$29.29 & 64.01$\pm{}$47.40 & 63.11$\pm{}$36.03 \\
\bottomrule
\end{tabular}
\end{table}

\begin{table}
\caption{The accuracy of predicting class of rating of ROUGE-L Recall scores on all non-adversarial questions. The class of a rating determined via quantizing ratings into bins defined by ranges, class 1 corresponds to scores of [0-33), class 2 to scores of [33-66), and class 3 to scores of [66-100].
The first column reports the accuracy over all the three classes, the subsequent columns report accuracies for individual classes from the lowest ROUGE class (``accuracy class 1'') to the highest class (``accuracy class 3'').}
\label{tab:correlations}
\centering
\begin{tabular}{lllll}
\toprule
 & overall accuracy & accuracy class 1 & accuracy class 2 & accuracy class 3 \\
rating by &  &  &  &  \\
\midrule
Gemma-3-27b-it & 0.27$\pm{}$0.21 & 0.28$\pm{}$0.38 & 0.06$\pm{}$0.16 & 0.19$\pm{}$0.22 \\
Mistral-7b & 0.36$\pm{}$0.22 & 0.52$\pm{}$0.38 & 0.04$\pm{}$0.12 & 0.18$\pm{}$0.24 \\
baseline GPT-4.1-nano & 0.33$\pm{}$0.21 & 0.47$\pm{}$0.41 & 0.07$\pm{}$0.14 & 0.17$\pm{}$0.23 \\
\bottomrule
\end{tabular}
\end{table}

\begin{table}
\caption{The accuracy for predicting class of rating of ROUGE-L Recall on expert-made questions.
For details see the caption of \cref{tab:correlations}.
}
\label{tab:correlationsgold}
\centering
\begin{tabular}{lllll}
\toprule
 & overall accuracy & accuracy class 1 & accuracy class 2 & accuracy class 3 \\
rating by &  &  &  &  \\
\midrule
Gemma-3-27b-it & 0.46$\pm{}$0.27 & 0.01$\pm{}$0.06 & 0.04$\pm{}$0.17 & 0.40$\pm{}$0.30 \\
Mistral-7b & 0.32$\pm{}$0.29 & 0.10$\pm{}$0.24 & 0.01$\pm{}$0.07 & 0.21$\pm{}$0.27 \\
baseline GPT-4.1-nano & 0.46$\pm{}$0.28 & 0.09$\pm{}$0.18 & 0.04$\pm{}$0.17 & 0.32$\pm{}$0.34 \\
\bottomrule
\end{tabular}
\end{table}

\begin{table}
\caption{The accuracy for predicting class of rating of ROUGE-L Recall on questions created by the procedure described
in \cref{demagog-questions-construction} from the Demagog database.
For details see the caption of \cref{tab:correlations}.
}
\label{tab:correlationsstatement}
\centering
\begin{tabular}{lllll}
\toprule
 & overall accuracy & accuracy class 1 & accuracy class 2 & accuracy class 3 \\
rating by &  &  &  &  \\
\midrule
Gemma-3-27b-it & 0.50$\pm{}$0.26 & 0.01$\pm{}$0.07 & 0.06$\pm{}$0.19 & 0.43$\pm{}$0.31 \\
Mistral-7b & 0.26$\pm{}$0.26 & 0.06$\pm{}$0.13 & 0.01$\pm{}$0.09 & 0.19$\pm{}$0.27 \\
baseline GPT-4.1-nano & 0.50$\pm{}$0.28 & 0.10$\pm{}$0.18 & 0.06$\pm{}$0.19 & 0.34$\pm{}$0.36 \\
\bottomrule
\end{tabular}
\end{table}

\begin{table}
\caption{The accuracy for predicting class of rating of ROUGE-L Recall. The questions include ones made adversarial by corrupting text or swapping material, the reference answers are kept unchanged. For details see the caption of \cref{tab:correlations}.}
\label{tab:correlationsadversarial}
\centering
\begin{tabular}{lllll}
\toprule
 & overall accuracy & accuracy class 1 & accuracy class 2 & accuracy class 3 \\
rating by &  &  &  &  \\
\midrule
Gemma-3-27b-it & 0.39$\pm{}$0.24 & 0.42$\pm{}$0.37 & 0.04$\pm{}$0.14 & 0.18$\pm{}$0.22 \\
Mistral-7b & 0.39$\pm{}$0.21 & 0.51$\pm{}$0.36 & 0.03$\pm{}$0.13 & 0.22$\pm{}$0.24 \\
baseline GPT-4.1-nano & 0.42$\pm{}$0.25 & 0.51$\pm{}$0.35 & 0.05$\pm{}$0.13 & 0.11$\pm{}$0.20 \\
\bottomrule
\end{tabular}
\end{table}